# Improved Super-Resolution Convolution Neural Network for Large Images

Rong Song    Jason Wang

**Abstract---** Single image super-resolution (SISR) is a very popular topic nowadays, which has both research value and practical value. In daily life, we crop a large image into sub-images to do super-resolution and then merge them together. Although convolution neural network performs very well in research field, if we use it to do super-resolution, we can easily observe cutting lines from merged pictures. To address these problems, in this paper, we propose a refined architecture of SRCNN with 'Symmetric Padding', 'Random Learning' and 'Residual learning'. Moreover, we have done a lot of experiments to prove our model performs best among a lot of the state-of-art methods.

**Index term----Super-resolution, cutting lines, image quality, random selection**

# 1 Introduction

Single image super-resolution (SISR) [1], aiming at recovering a high-resolution image (HR) from its low-resolution counterpart, is a research topic of common interest in computer vision. Generally speaking, image super-resolution (SR) is a pending question, and various algorithms have been developed over these decades. Existing SR algorithms mainly include three parts: interpolation-based SR [2], reconstruction-based SR [3] and learn-based SR [4]. Among them, learn-based SR algorithms have drawn the highest attention and achieve the state-of-art performance along with the development of machine learning. The key to learn-based SR algorithms is to find a nonlinear mapping between low-resolution images and high-resolution images.

Inspired by the recent successes  in image processing achieved by the convolutional neural network (CNNs) [12][13][14], Dong et al. have demonstrated a novel method named SRCNN [5] which outperforms many classical algorithms merely using a light architecture. Next, considering its limited training speed, Dong et al. further designed a Fast-Super-Resolution Convolutional Neural Network (FSRCNN) [6], in which they explored a more efficient network structure to achieve high running speed without the loss of restoration quality, thus making it possible to realize real-time video SR. Besides those optimizations on SRCNN, it is still very difficult to further improve the image quality with a shallow network. Inspired by the Visual Geometry Group net (VGGNet) [7], Kim deepened its CNN architecture for SR to 20 layers and proposed a very deep network (VDSR). To improve its training efficiency, in addition, [7] innovatively brought residual learning into the deep net architecture.

However, there are still some flaws in even the most advanced SR network structure. Despite those optimizations on CNN architecture, training speed and training time still be a very hard issue to cope with. Moreover, existing grids and cutting lines in the merged images after processing is the biggest obstacle to actual production, as illustrated in *fig. 1.*

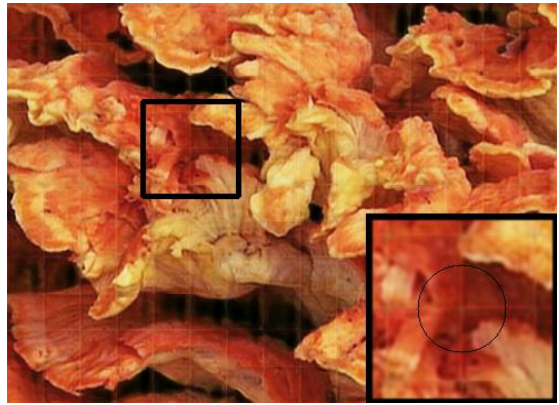

Figure 1:    A merged image resulted from traditional SRCNN. It is easily to observe cutting lines between neighboring sub images.

To address these problems, we focus on SRCNN so as to explore the merits of CNN, such as the high speed of computing, high accuracy by end-to-end training and so on [8][10]. The *main contribution* of this paper is summarized in the following:

1) Add 'Symmetric Padding' before each convolutional layer, avoiding introducing useless information to images and keep the input and output size the same.

2) Invent 'Random Learning' to speed up our training efficiency.
3) Apply 'Residual Learning' to traditional SRCNN, making the architecture deeper than traditional SRCNN.

These improvements enable SRCNN to have higher performance for handling real-world problems.

Experimentally, to validate the effectiveness of our method, we test it on g2.x2large GPU from EC2 of AWS. We trained our model for several hours and compare it with traditional SRCNN on Set5 and Set14. The results prove that our model has the best PSNR compared with others. Besides, via controlling variables, we prove the learning efficiency of 'random learning' is indeed higher than the ordinary one.

This paper is organized as followed: Section II makes a brief introduction on SRCNN, residual learning, and padding, which are closely related to our work. Section III shows the architecture of our neural network specifically. Section IV lists a lot of experiments and analyses we have done. Section V is the conclusion of our conclusion.

## 2 Related work

### 2.1 Super-Resolution Convolutional Neural Network

The first method of realizing SISR using deep learning can be traced back to [5], where three layers of CNN named SRCNN is designed to solve the problem efficiently. This simple but light structure can be divided into three parts, which are the 'Patch extraction layer', 'Non-linear mapping' and 'Reconstruction'. They directly consider a convolutional neural network which is an end-to-end mapping function between photos with different resolution. The reason why this is a surprising invention is that they intentionally designed with simplicity in mind, and yet provides great accuracy and speed even compared to state-of-art example-based methods.

**2.2 Residual Learning**

Network depth is of crucial importance [15]. People can get excellent outcomes using very deep structures [16][18] to train models because a deep architecture of neural network tends to perform better than a shallow one. However, the training speed of a deep model is likely to be low because of the vanishing/exploding gradients. To solve this problem, He et al. [10] developed a new structure called deep residual learning, which adds a shortcut from the formal layer to the trained layer. The network structure largely solves the vanishing/exploding gradient and ensure that a deeper model should produce no higher training error than its shallower counterpart [12].

**2.3 Padding**

In order to make the input and output size the same size, one classic solution is to add zero padding to the edges of the image [7]. In every layer, the neural network will add some zero padding to make the output have a same size of the input. However, the zero-padding method may bring some useless information and even jeopardize the reconstruction of images [19]. Luo et al. [19] invented a new way to deal with this problem. They calculate the difference in the size of input and output according to convolutional kernel and stride of each convolutional layer and add some symmetric paddings to model's original input images. With this technique, the network can avoid introducing useless information into the input image and can make the input and output images in the same size.

# 3 Proposed Methodology

### 3.1 Framework

An overview of our network is portrayed in *Fig. 2*

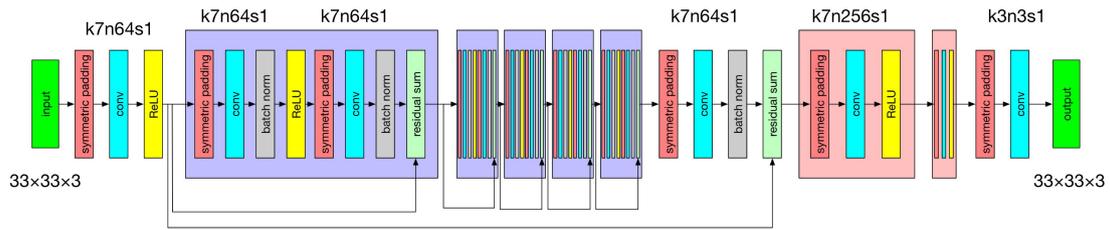

Figure 2: The architecture of our SRCNN, a refined structure from the generator from SRGAN [9], with kernel size (k), feature maps (n) and stride(s). Training batch is 8.

We will show the details of our distinctive techniques: random training and symmetric padding. And then we will explain our architecture specifically.

**3.2 Random Training**

As we mentioned, it is not appropriate to directly regard all the small images as input. Assuming input images' sizes are about (400,400,3). If we crop them into (33,33,3), we will get about 144 sub-images from a single picture, indicating that for a single image, the neural network will train about 18 steps which makes training process take a long time.

Our solution is to reduce the number of input images. Firstly, we divide a large picture into $n$ small pieces and rearrange them into $n/b$ batches where $b$ is the size of each batch. Next, we will randomly select $k$ (from 0 to $n/8b$) batches among them *(Fig. 3)*. By doing this, we can cut down the number of input samples to about one-sixteenth of the original input amount.

We originally thought that reducing the number of input samples would result in poor results. Interestingly, the opposite is true. To show random learning is more efficient, we will compare it with a traditional one in the experimental part.

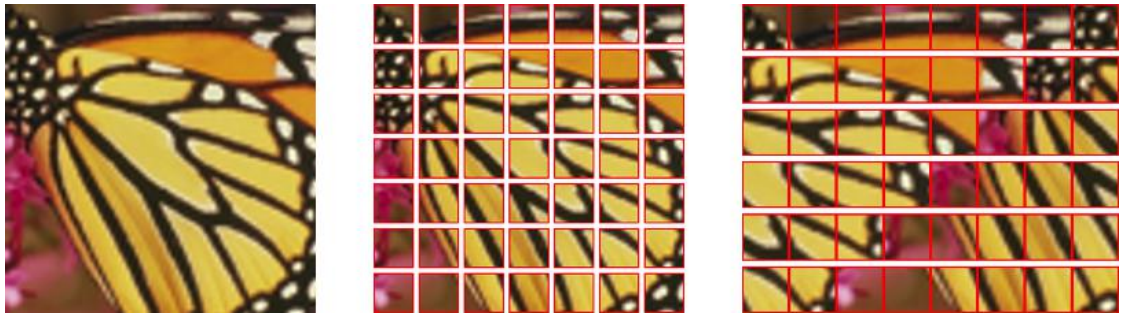

Figure 3: The left one is our original input. We will cut it into 49 pieces of blocks with the size of 33x33 pixels. We rearrange sub images from start to finish in order and get series of batched from the left picture, where n is 6 and b is 8. The shape of array is [6,8,33,33,3]

**3.3 Symmetric Padding**

Big images in the actual process need to be cut into small pieces due to limited RAM space and other hardware requirements. In order to make the input images and the output images the same size, traditional SRCNN apply zero padding before each convolutional layer. However, while testing traditional SRCNN, we found that some unavoidable lines occur in merged pictures which will seriously affect the quality of human eye observation. What's more, this flaw even makes it impossible to put SRCNN into daily use. For example, in Satellite Imagery [19], after using the traditional SRCNN, the local part can be destroyed greatly, thus will affect the whole analyze.

In order to avoid introducing useless information, another strategy [5] does not add padding at the edges of image layers before convolution, which results in output size is smaller than the input size.

To solve these problems, we apply 'symmetric padding' to our model. Our strategy is to an add symmetric padding before each convolution layer.

Let us consider one-dimensional convolutional layer, whose size is $(h, w)$, kernel size is $(k_1, k_2, 1)$ and stride is $(s_1, s_2)$. We can easily calculate our desired symmetric padding, whose size is $(s_1 h + k_1 - 1, s_2 w + k_2 - 1)$, which guarantee the output size is the same as the origin size after this convolution layer.

When we have known the desired size of input images, we can fill the margin information symmetrically. A comparison between zero padding and 'mirror padding' is shown in *Fig. 4*.

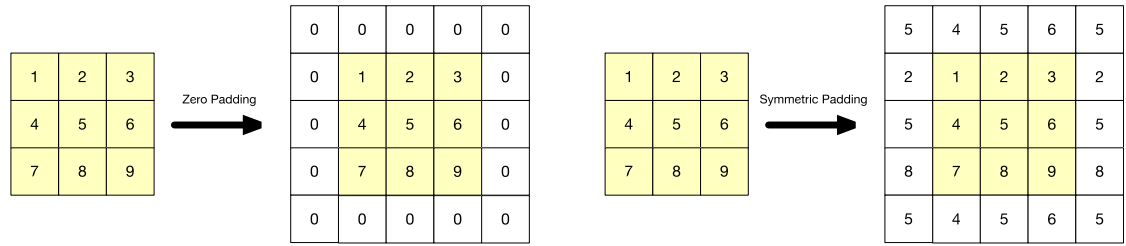

Figure 4:   Left one is zero padding, and right one is symmetric padding sample. Numbers inside grids are the value of neurons.

After adding symmetric padding, pixels at the edges free from junk information and gain more useful information from their neighboring pixels. As a result, symmetric padding not only keeps input and output in the same size but makes pixels at the edges clearer.

**3.4 Residual learning**

Many other tasks [20][21] about computer visual have greatly benefited from very deep models, but an impede is vanishing/exploding gradients [10]. To improve the performance of our model, what we do is to deepen the layer in non-linear mapping with residual learning.

Our architecture is inspired by the generator from SRGAN [9] with residual learning, which turns out works pretty well. In this paper, we define a residual block as:

$$x^{'} = R(x) + x$$

Here we assume that $x$ represents the input of a residual block and $x'$ represents the output of that residual block. Each residual block $R_i$ has two convolutional layers, $R_i = W_{2i}P(\sigma(W_{2i-1}P(x)))$ in which denotes $P$ symmetric padding layer shown before, $\sigma$ stands for activation function ReLU [22] and $W_i$ is the convolutional layer $i$ with weight $w_i$.

**3.5 Nonlinear Mapping**

Firstly, we arrange a convolutional layer where kernel size is (33,33,64) and its dimension is 3. After that, the size of the output is (33,33,64)

Then, there are 5 residual blocks at the first phase of non-linear mapping, each block including two convolution layers in which 64-dimensional feature filter is (7,7,64), the stride is 1. There is a convolution layer whose filter's size is the same as that in residual blocks with a shortcut connection behind that. Next, there are two convolution layers where 64-dimensional feature filter's size is (7,7,256). The size of output is (33,33,256).

Finally, there is only one convolution layer where the size of 256-dimensional kernel size is (33,33,3) and stride of it is 1. Through nonlinear mapping, we finally get our output image with the same size with input image.

**3.6 Loss Function**

Mean squared error (MSE) is often used as a criterion to measure the distance between output and ground truth. For image $y_i$ is real super-resolution image and $y'_i$ is our generated super-resolution image, the MSE is (1).

$$MSE = \frac{1}{size\ of\ image}\|y_i - y'_i\|^2 \quad (1)$$

In the field of Super Resolution, one of the most important indicators is Peak Signal to Noise Ratio (PSNR) which is calculated by (2) where n is the number of bits of a pixel. Obviously, the lower MSE is, the higher PSNR is.

$$PSNR = 10 log_{10} \frac{(2^n-1)^2}{MSE} \quad (2)$$

During our training, our loss function is (3) where N is the batch size, which is actually proportional to MSE. If we try to minimize our loss, PSNR will increase theoretically.

$$Loss = \frac{1}{2N} \sum_{i=1}^{N} \left\| y_i - y_i' \right\|^2 \quad (3)$$

## 4 Experiment and Analyze

### 4.1 Dataset and Experimental Environment

For a fair comparison with previous work, we use the ImageNet to train our model. Because different pictures may get the outcomes with quite different PSNR and SSIM, we test our modified algorithm using the same datasets in [5]. All the experiments are done on the g2.x2large GPU from EC2 of AWS.

### 4.2 Results

In *Fig. 5*, we can find that the longer time we train the deeper lines will occur in the picture, which is caused by the lost specific details that are filling with zeros. After we insert a function to do the symmetric padding rather than just using zero padding or discarding the edges, lines become harder to find. To make it more explicitly, we display two sets of images using two different kinds of methods with the same training time in *Fig. 6* and *Fig. 7*.

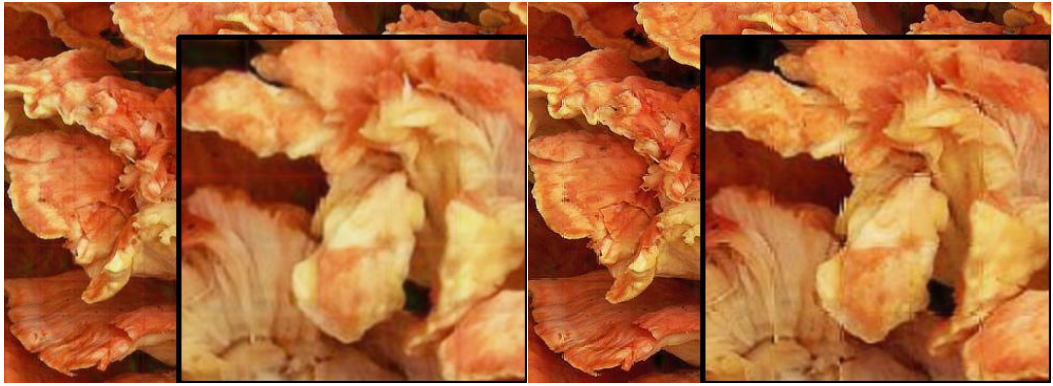

Figure 5: With the more epochs (from the left one to the right one), we can find the cutting lines are deeper and more broken in the magnified parts

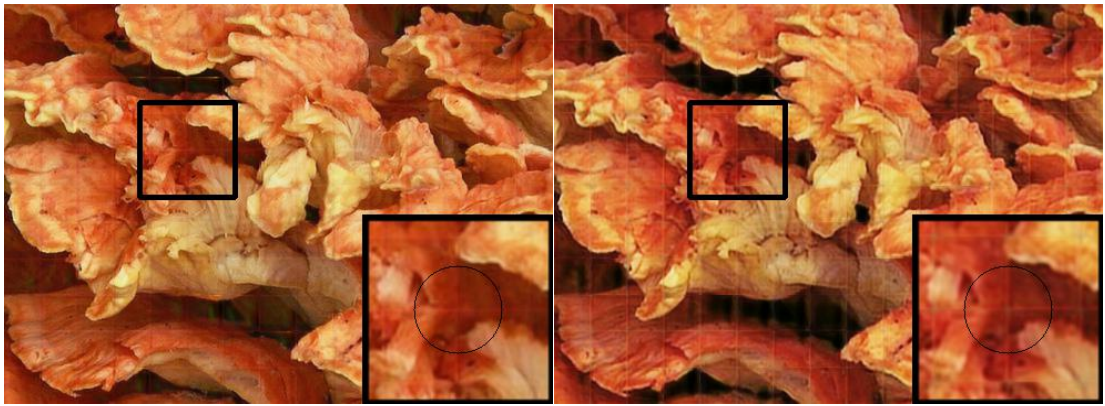

Figure 6: The pictures shows outcomes using traditional merging method. It can be see that with the training time increasing, the cutting lines is becoming more obvious even though the whole quality of picture is increasing.

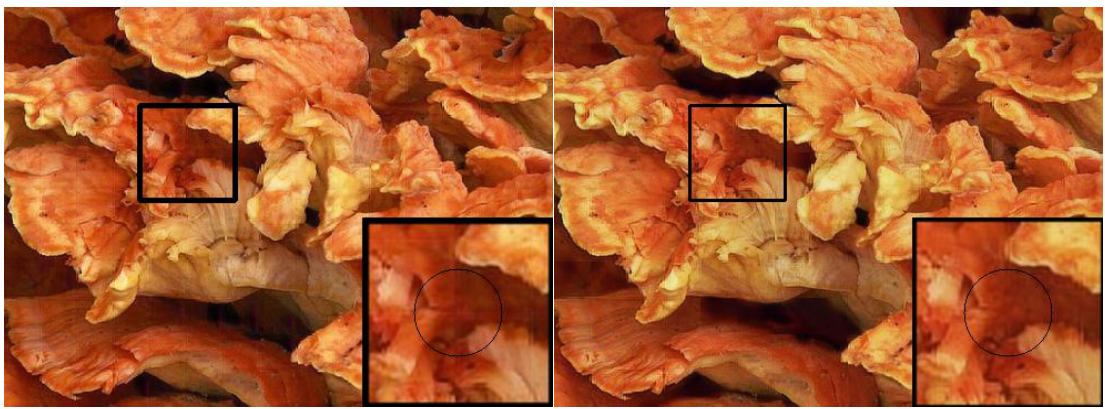

Figure 7: The pictures shows outcomes using our novel padding filling method. It can obviously find the lines are dramatically declined compare to PIC2 with the same training time.

## 4.3 Analysis

Adding mirror padding can save the lost details that used to be discarded due to the convolution kernel, thus making the model be able to learn how to reconstruct the splitting lines. To prove our method can also generally applicable to most situations, we use two test sets to calculate SSIM and PSNR in *Table. 1*. The experiment shows our method makes it become possible to put the picture into actual production after reconstruction.

| PSNR  |    | bicubic   | SRCNN_zero | SRCNN_symmetric | ours      |
|-------|----|-----------|------------|-----------------|-----------|
| set5  | *3 | 30.318883 | 31.646021  | 31.524616       | 32.072755 |
| set14 | *3 | 27.439712 | 28.358123  | 28.398823       | 28.457813 |
| SSIM  |    | bicubic   | SRCNN_zero | SRCNN_symmetric | Ours      |
| set5  | *3 | 0.837259  | 0.863555   | 0.866047        | 0.873555  |
| set14 | *3 | 0.737820  | 0.764741   | 0.770077        | 0.783175  |

Table 1: Comparison of results for using bicubic interpolation, SRCNN using zero padding, SRCNN using symmetric padding and ours

We use the classic shuffle methods to make sure to intermingle all the data to make the network learn different information rather than fixed learning sequence during every epoch. Because 33×33 pieces can be the best training size [5][7], original pictures will be cut into hundreds of thousands of small pieces. Next, we not only cut the big picture into small pieces but also choose numbers and gradation from the whole batch randomly to make net learn more general knowledge. (*fig. 3*)

To prove our improvement, we do two sets of experiments. We set the same training time at 30000s totally to compare the different performances. It can be seen in *Fig. 8* that by using our "random learning method", the time used each epoch is much shorter and unbalanced than the traditional methods. When it comes to PSNR, at the same time, the PSNR calculated from our method is quite higher than the traditional one, which suggests that randomly learning is better than normal learning. From this perspective, the training process can save a lot of time.

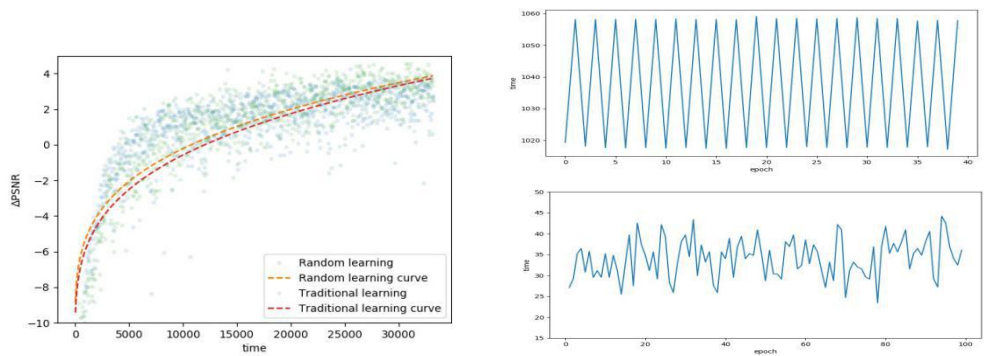

Figure 8  This left picture shows that at the first 30000s the PSNR of random learning is obviously higher than traditional learning, showing that "random learning" can accelerate the training speed. The right two pictures show the usage of time each epoch

# 5 Conclusion

In this paper, we have proposed a super-resolution neural network using symmetric padding to handle the situation where the image is too large. To improve out training speed, we also apply residual-learning and random learning to our model. We have demonstrated that our model can eliminate the border between two different sub-image and our model can be trained faster than other counterparts. Because all current super-resolution neural networks can't handle too large pictures, we believe our model is very meaningful.